\documentclass{article}
\usepackage[utf8]{inputenc}
\usepackage{spconf,amsmath,graphicx,hyperref}
\usepackage{graphics,graphicx,color,epsfig,subcaption}
\usepackage{amsfonts}
\usepackage{amsthm,amssymb}
\usepackage{nicefrac}       
\usepackage{xcolor}
\usepackage{multirow}
\usepackage{booktabs}       
\usepackage[table]{xcolor}
\usepackage{colortbl}
\usepackage{tikz}
\usetikzlibrary{automata, positioning, arrows,shapes.multipart,fit}

\usepackage{algorithmic}
\usepackage{algorithm}

\usepackage[utf8]{inputenc}
\usepackage{devanagari}  

\definecolor{rank1}{RGB}{230,190,255} 
\definecolor{rank2}{RGB}{200,210,255} 
\definecolor{rank3}{RGB}{140,220,255} 
\definecolor{rank4}{RGB}{200,255,200} 
\definecolor{rank5}{RGB}{255,255,180} 
\definecolor{rank6}{RGB}{255,210,170} 
\definecolor{rank7}{RGB}{255,180,180} 

\title{Structured Sparsity and Weight-adaptive Pruning for Memory and Compute efficient Whisper models}
\name{Prasenjit K Mudi$^{1}$, Anshi Sachan$^{2*}$, Dahlia Devapriya$^{1*}$, Sheetal Kalyani$^{1}$ \thanks{$^{*}$Authors contributed equally; $^{2}$anshijio123@gmail.com\\$^{1}$ \{ee21d057@smail, ee22d003@smail, skalyani@ee\}.iitm.ac.in}}
\address{$^{1}$Indian Institute of Technology Madras, India\\ $^{2}$National Institute of Technology Karnataka, Surathkal\\
}
\begin{document}
%
\maketitle
\begin{abstract}
Whisper models have achieved remarkable progress in speech recognition; yet their large size remains a bottleneck for deployment on resource-constrained edge devices. This paper proposes a framework to design fine-tuned variants of Whisper which address the above problem. Structured sparsity is enforced via the Sparse Group LASSO penalty as a loss regularizer, to reduce the number of FLOating Point operations (FLOPs). Further, a weight statistics aware pruning algorithm is proposed. We also design our custom text normalizer for WER evaluation. On Common Voice 11.0 Hindi dataset, we obtain, without degrading WER, (a) 35.4\% reduction in model parameters, 14.25\% lower memory consumption and 18.5\% fewer FLOPs on Whisper-small, and (b) 31\% reduction in model parameters, 15.29\% lower memory consumption and 16.95\% fewer FLOPs on Whisper-medium; and, (c) substantially outperform the state-of-the-art Iterative Magnitude Pruning based method by pruning 18.7\% more parameters along with a 12.31 reduction in WER.
\end{abstract}
\section{INTRODUCTION}\label{sec:intro}
There has been active research on fine-tuning large pre-trained models like Whisper \cite{radford2023robust} for Automatic Speech Recognition (ASR) in different low-resource languages, such as Amharic \cite{gete2025whispering}, Nepali \cite{rijal2024whisper}, Swiss German \cite{timmel2024fine}, Afrikaans, Belarusian, Icelandic, Kazakh, Marathi, Nepali, and Swahili \cite{liu2024exploration}. The IndicWhisper model in \cite{kumar2023towards} fine-tunes Whisper-medium on $10,700$ hours of data belonging to $12$ Indian languages, including Hindi. In edge applications limited by compute and memory, one would typically like to prune these models.

The pruning of neural networks has been studied in literature, starting from the landmark paper\cite{lecun1989optimal} which demonstrated pruning without loss of accuracy for a simple neural network based on second order loss derivatives. The authors of \cite{lth} demonstrated that through iterative magnitude pruning (IMP), one can recover a subnetwork that reaches an accuracy comparable to that of the full network, known as the lottery ticket hypothesis (LTH). Based on this hypothesis, \cite{audiolottery} pruned a transformer model with IMP. Following this, the authors of \cite{pepsi} applied the pruning methodology of LTH to the Whisper model and obtained results comparable to the unpruned model. However, the existing works have three major limitations: (1) the pruning methods are iterative\cite{audiolottery,pepsi}, (2) they require additional adapters plugged into the model layers, increasing the inference time complexity\cite{pepsi}, and, (3) pruning with IMP does not reduce the number of FLOPs as the pruning is unstructured.

In order to address these shortcomings, this paper proposes a fine-tuning and pruning method, called TSPAR (fine\textbf{T}uning for \textbf{S}tructured s\textbf{P}arsity and \textbf{A}dapti-ve p\textbf{R}uning), which is applied on the Whisper-small and Whisper-medium models. In pruning literature, loss regularization has traditionally been used to sparsify the network during training \cite{lecun1989optimal,han2015learning,alvarez2016learning,nayak2020green}. However, imposing a penalty on the weight magnitudes only imposes unstructured sparsity, which does not reduce the number of FLOating Point operations (FLOPs). Using the Sparse Group LASSO (SGL)\cite{simon2013sparse} penalty function to regularize loss, one can achieve structured sparsity wherever it exists and unstructured sparsity elsewhere, which saves both FLOPs and memory utilization. We employ SGL regularized fine-tuning on Whisper, and prune according to the weight statistics of the regularized model, achieving comparable results to the full model. Ours is a non-iterative approach, which prunes up to $51.5\%$ parameters in one shot with a considerable reduction in FLOPs, without requiring any additional adapters.

The WER metric has traditionally been used to evaluate ASR models. Existing normalizers such as Whisper’s default “Non-English” normalizer\cite{radford2023robust} or the Hindi normalizer from Indic NLP library\footnote{https://indic-nlp-library.readthedocs.io/en/latest/} are not ideal: Whisper’s normalization is overly aggressive and discards meaningful diacritics, while Indic normalization preserves all constructs. 
This makes WER values across languages unfair; Hindi ends up with a more misleading WER than less phonemically complex languages like English \cite{ohala1992nasals}. 
We propose a more balanced text normalization for Hindi, and adopt it in our WER computation. In summary, our main contributions are threefold: (1) imposing structured sparsity on fine-tuned Whisper models using the SGL penalty function, thereby reducing both memory footprint and FLOPs; (2) proposing a weight-statistics–driven pruning strategy that adapts dynamically to model parameters; and (3) designing a custom Hindi text normalizer for fair WER evaluation.

\section{PROPOSED METHOD}\label{sec:method}
\subsection{Regularized training for structured sparsity}\label{sec:loss_reg}
Loss regularization techniques such as weight decay or $L_1$ norm minimization involves adding a penalty term on the magnitude of the weights to regularize the cross entropy function $\mathcal{L}_{CE}$. This falls in the category of regularized loss minimization, while just using $\mathcal{L}_{CE}$ is empirical risk minimization. The sparsity induced by the $L_1$ mechanism is unstructured, as the weights being driven to zero are randomly spread across the weight matrices. Pruning these parameters reduces memory requirement, but does not reduce FLOPs.
In the context of statistical literature, Sparse Group LASSO (SGL) \cite{simon2013sparse} was introduced where, an additional grouped loss regularization  imposes structured sparsity on a group of weights as a whole. We exploit this SGL penalty function in our work to drive certain columns of weight matrices to zero. 
Let $\mathbf{W}_{jl}$ be the $j^{th}$ weight matrix of the $l^{th}$ layer, with a total of $J_l$ matrices for $1\leq l \leq L$ layers. Let $\mathbf{W}_{jl}^i$ denote the $i^{th}$ column of $\mathbf{W}_{jl}$ and $I_{J_l}$ denote the number of columns of each matrix $\mathbf{W}_{jl}$. Each such column $\mathbf{W}_{jl}^i$ represents the set of outgoing connections from a neuron in layer $j$ to every neuron in the next layer $j+1$. By imposing column-wise structured sparsity, the entire column can be pruned. These pruned columns are excluded from matrix multiplication, resulting in the reduction of FLOPs. Hence, the total loss is defined as,
\begin{equation}\label{eqn:loss}
    \mathcal{L} = \mathcal{L}_{CE} + \lambda_1 (\mathcal{L}_{L_{1}}^{E}) + \lambda_2 (\mathcal{L}_{L_{1}}^{D}) + \lambda_3 (\mathcal{L}_{L_{2}}^{E}) + \lambda_4 (\mathcal{L}_{L_{2}}^{D})
\end{equation}
\begin{equation}\label{eqn:l1_loss}
    \mathcal{L}_{L_{1}}^{E,D} =  \lambda_{1,2} \sum_{l=1}^L\left(\sum_{j=1}^{J_l} \left(\left\|\mathbf{W}_{jl}\right\|_1\right)\right)
\end{equation}
\begin{equation}\label{eqn:l2_loss}
    \begin{aligned}
        \mathcal{L}_{L_{2}}^{E,D} = \lambda_{3,4} \sum_{l=1}^L\bigg(\sum_{i=1}^{I_1}\left\|\tilde{\mathbf{W}}_{1 l}^i\right\|_2 +\cdots + \sum_{i=1}^{I_{J_l}}\left\|\tilde{\mathbf{W}}_{J_{l} l}^i\right\|_2\bigg)
    \end{aligned}
\end{equation}
where superscripts E and D on $\mathcal{L}$ denote the encoder and decoder, while subscripts $L_{1}$ and $L_{2}$ denote the regularization terms that induce unstructured and structured sparsity, respectively. 
\begin{algorithm}
\caption{Pruning}
\label{alg: pruning}
\begin{algorithmic}[1]   
    \REQUIRE weight matrix $W$
    \STATE set $\theta_{0} = 0.1 \sigma_{c}$, $\theta_{c}=0.9$
    \STATE set $\theta_{w} \hspace{6mm} \text{as per Table \ref{tab: pruning_FC}}
    $
    \FOR{each column $C$ in weight matrix $W$}
        \IF{$\hat{S}_{c}(w;\theta_{0}) \geq \theta_{c}$}
            \STATE reset $C = $ \textbf{0}
        \ENDIF
    \ENDFOR
    \FOR{each unpruned column in weight matrix $W$}
        \IF{($\hat{S}_{c}(w;\theta_{0}) < \theta_{c}$ AND $|w| < \theta_{w}$)}
            \STATE reset $w = 0$
        \ENDIF
    \ENDFOR
\end{algorithmic}
\end{algorithm}
\begin{figure*}[!t]
    \centering
    \begin{subfigure}[b]{0.19\textwidth}
        \centering
        \includegraphics[width=\textwidth]{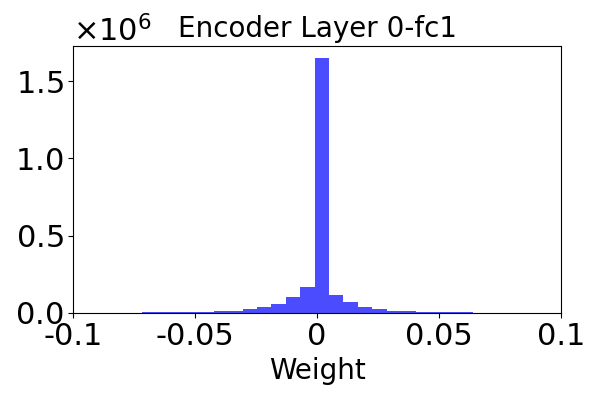}
        \caption{highly spiked}
        \label{fig:fig1}
    \end{subfigure}
    \hfill
    \begin{subfigure}[b]{0.19\textwidth}
        \centering
        \includegraphics[width=\textwidth]{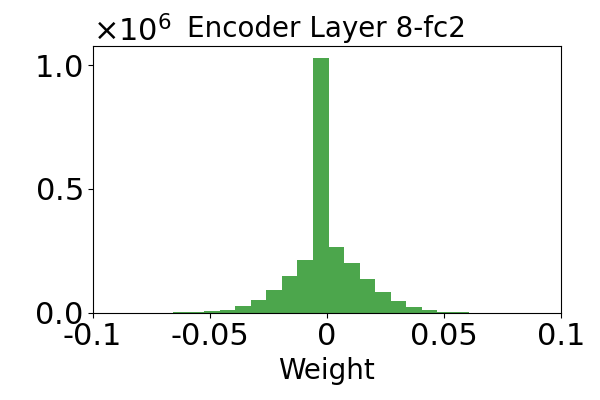}
        \caption{narrow spread}
        \label{fig:fig2}
    \end{subfigure}
    \hfill
    \begin{subfigure}[b]{0.19\textwidth}
        \centering
        \includegraphics[width=\textwidth]{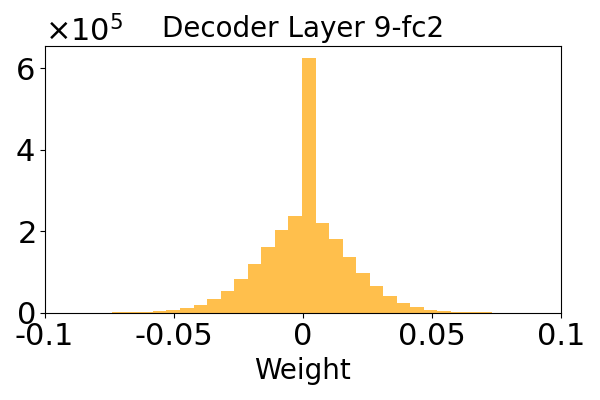}
        \caption{wide spread}
        \label{fig:fig3}
    \end{subfigure}
    \hfill
    \begin{subfigure}[b]{0.19\textwidth}
        \centering
        \includegraphics[width=\textwidth]{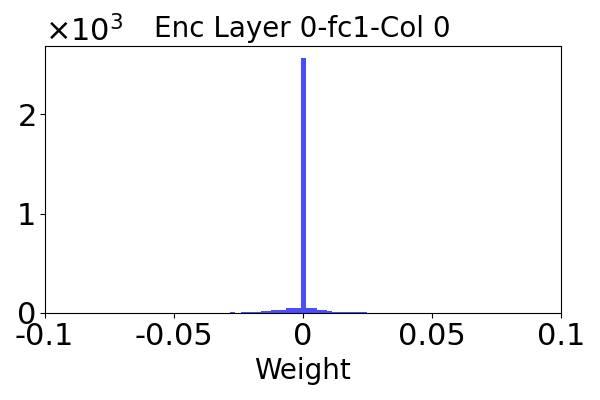}
        \caption{pruned column}
        \label{fig:fig4}
    \end{subfigure}
    \hfill
    \begin{subfigure}[b]{0.19\textwidth}
        \centering
        \includegraphics[width=\textwidth]{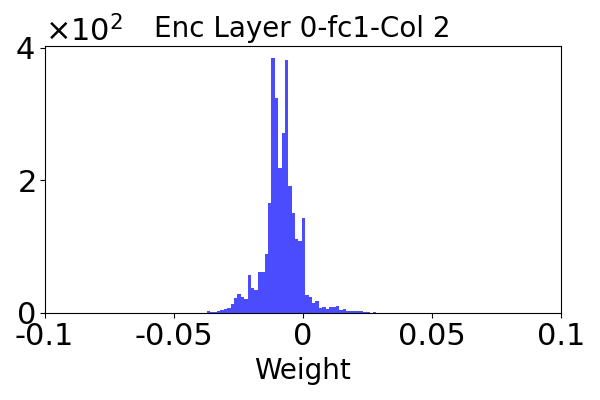}
        \caption{unpruned column}
        \label{fig:fig5}
    \end{subfigure}
    \caption{Different types of weight distributions}
    \label{fig:five_figs}
\end{figure*}
\subsection{Weight adaptive pruning}
We propose a two-pass pruning algorithm, which is given in Algorithm \ref{alg: pruning}. The structured $L_2$ loss regularization drives entire columns to zero, and hence in the first pass (lines 3-7) we prune columns which are approximately sparse. The columns with high density of ``almost zero weights" are called "approximate sparse columns". A weight, $w$ is defined as an ``almost zero weight" if $|w| < \theta_{0}$ for some positive constant $\theta_{0}$. A column $C$ of a weight matrix is defined as ``approximately sparse" if the approximate sparsity of the column, $\hat{S}_{c}$ is greater than a positive threshold $\theta_{\hat{S}_c}$. Approximate sparsity is defined as $\hat{S}_{c}(w;\theta_{0}) = \frac{\mathbf{1}_{|w|<\theta_{0}} (C)}{|C|}$ where, $\mathbf{1}(.)$ is an indicator function, and $|C|$ is the length of column $C$. During the second pass of pruning (lines 8–12), we prune individual weights that have been driven to zero by the $L_1$ loss regularization with a threshold $\theta_w$.
While pruning thresholds are typically tuned by trial and error, we propose to analyze the weight statistics of the model and derive pruning thresholds $\theta_{0}, \theta_{c}$, and $\theta_{w}$. The detailed analysis is provided in Section \ref{sec:wt_stat_analysis}. 
\section{ANALYSIS OF WEIGHT STATISTICS}\label{sec:wt_stat_analysis}
In this section, we analyze the weight statistics of three fine-tuned models when regularization is applied on: (a) only fully connected (FC) layers of Whisper-small, (b) all layers of Whisper-small, and (c) only FC layers of Whisper-medium. These models are referred to as Net-1, Net-2, and Net-3, respectively. In all models, the mean $\mu$ and standard deviation $\sigma$ of weights are in the order of $10^{-4}$ and $10^{-2}$, respectively, in all layers (except LayerNorm in Net-2). We observe three main types of weight distributions, namely (A) highly spiked (Fig.\ref{fig:fig1}), (B) narrowly spread (Fig.\ref{fig:fig2}), and (C) widely spread (Fig.\ref{fig:fig3}) distributions around mean. In highly spiked distributions (type A), the weights between the $25^{th}$ and $75^{th}$ percentiles, denoted by $Q_2, Q_3$, are in the order of $10^{-11}$. This implies that $50\%$ of the weights are very close to zero, and hence, can be pruned immediately. However, we prune all weights between $\mu - 0.1 \sigma$ and $\mu + 0.1 \sigma$ in order to prune more weights than those in the interquartile range. So, we set the threshold function as $T_{1} = 0.1 \sigma$. In narrowly spread distributions (type B), $Q_2, Q_3$ are at $\sim (\mu \pm 0.3\sigma)$. To prune slightly more weights than those in the inter-quartile range, the threshold function is set as $T_{2}=\max\{|Q_2|, |Q_3|\}$. The widely spread distributions (type C), have $Q_2, Q_3$ locations at $\sim (\mu \pm 0.67 \sigma)$. We can not apply threshold to be $\max\{|Q_2|, |Q_3|\}$ as it will  prune too many weights. Hence, we use the widely adopted threshold function, $T_{3}=\eta W_{max}$ where $W_{max}$ is the maximum absolute weight of matrix $W$ and $\eta > 0$ is a hyperparameter \cite{nayak2020green}.

We analyze the weight distributions of the columns of weight matrices in different layers. The columns that follow highly spiked distributions (type A), have weights very close to mean ($\sim 0$) i.e., almost zero weights. To identify these weights, $\theta_{0}$ is set as $0.1 \sigma_{c}$. To prune columns with 90\% approximate sparsity, $\theta_c$ is set as $0.9$ (see a sample pruned column weight distribution in Fig. \ref{fig:fig4}). The weights of the unpruned columns (a sample weight distribution shown in Fig. \ref{fig:fig5}) of a layer are further pruned in the second pass of pruning using a threshold $\theta_w$, which is set as $T_1, T_2$ or $T_3$, depending on whether the layer follows distribution type A, B, or C. We observe the following weight distributions in Net-1, 2, and 3.\\
\textbf{Net-1:} The first 6 FC layers of the encoder have type A distributions, remaining encoder layers follow type B, while the decoder layers follow type C distributions.\\
\textbf{Net-2:} There are 3 FC + 7 attention (ATT) + 1 convolution (CONV) layers in the encoder, and 14 ATT + 2 FC layers in the decoder; i.e., a total of 27 layers (denoted by $\mathcal{P}_1)$, which exhibit distributions of type A. The remaining layers follow distributions of type C. The LayerNorm layers of both the encoder and decoder have significantly larger weights ($\mu \sim 1.0, \sigma \sim 10^{-1}$) and hence, cannot be pruned.\\
\textbf{Net-3:} The first 5 layers of the encoder follow type A, while all the remaining layers of the encoder and decoder follow type B distribution. We apply the derived thresholds and provide results in the next section. 

\section{RESULTS AND DISCUSSION}\label{sec:results}
As Hindi is the third largest (609M total speakers) language in the world \footnote{https://www.ethnologue.com/insights/most-spoken-language/}, we consider the benchmark Common Voice 11.0 Hindi dataset for our experiments. Whisper-small and Whisper-medium are selected for our experiments as WERs of these models are substantially better than Whisper-tiny, base for Hindi data (see Appendix D.2.2, D.2.4 in \cite{radford2023robust}). Our aim is to design models for resource constrained edges, and thus we do not choose the large, large v2 versions of Whisper. For fine-tuning, the pre-trained Whisper is trained with 5000 steps (24.39 epochs) with training batch-size and evaluation batch-size of 8, AdamW optimizer, cosine learning rate scheduler, warmup ratio of 0.2, and weight decay of 0.01 \cite{sharma2025fine}. 
We set $\lambda_{1}=\lambda_{2}=10^{-5}$ (for $L_{1}$ loss) and $\lambda_{3}=\lambda_{4} =10^{-4}$ (for $L_{2}$ loss) by tuning independently.
\begin{table}[h]
\centering
\begin{tabular}{|l|l|l|c|c|}
\hline
\textbf{} & \textbf{Encoder} & \textbf{Decoder} & \textbf{Pruning} & \textbf{WER} \\
\hline
\multirow{3}{*}{\rotatebox[origin=c]{90}{Net-1}}
& $T_3$ & $T_3$ & 41.7\% & 26.24 \\
\cline{2-5}
& $T_2$ & $T_3$ & 44.9\% & 26.75 \\
\cline{2-5}
& $\theta_{w}^{1}$ & $T_3$ & 35.4\% & 24.84 \\
\hline
\multirow{4}{*}{\rotatebox[origin=c]{90}{Net-2}}
& $T_3$ & $T_3$ & 38.4\% & 38.18 \\
\cline{2-5}
& $\theta_{w}^{2}$ & $T_3$ & 32.2\% &  32.86 \\
\cline{2-5}
& $\theta_{w}^{3}$ & $T_3$ & 28.5\% & 34.35 \\
\cline{2-5}
& $\theta_{w}^{4}$ & $T_3$ & 32.7\% & 33.0 \\
\hline
\multirow{3}{*}{\rotatebox[origin=c]{90}{Net-3}}
& $T_3$ & $T_3$ & 31.0\% & 19.58 \\
\cline{2-5}
& $T_2$ & $T_3$ & 38.9\% & 21.15 \\
\cline{2-5}
& $\theta_{w}^{5}$ & $T_2$ & 51.5\% & 27.8 \\
\cline{2-5}
& \multicolumn{4}{|c|}{
\begin{minipage}{0.80\columnwidth}
\footnotesize
\[
\theta_{w}^{2} = 
\begin{cases}
0.1 \sigma & \text{;$\mathcal{P}_{1}$}\\
0.1 T_1 & \text{;Enc-ATT}\\
0.1 T_2 & \text{;Dec-ATT}\\
0.2 T_1 & \text{;Enc-FC}\\
T_2 & \text{;Dec-FC}\\
\end{cases}
\quad
\theta_{w}^{4} = 
\begin{cases}
0.1 \sigma & \text{;$\mathcal{P}_{1}$}\\
0.1 T_1 & \text{;Enc-ATT}\\
0.5 T_2 & \text{;Dec-ATT}\\
0.2 T_1 & \text{;Enc-FC}\\
T_2 & \text{;Dec-FC}\\
\end{cases}
\]
\end{minipage}
} \\
\hline
\multicolumn{3}{|c|}{Whisper-small with IMP\cite{pepsi}} & 16.7\% & 37.15 \\
\hline
\multicolumn{3}{|c|}{Whisper-medium with IMP\cite{pepsi}} & 20.29\% & 30.49 \\
\hline
\end{tabular}
\caption{Pruning experiments for different thresholds ($\theta_{w}$) : $\theta_{w}^{1} = \{ \max\{T_1, T_2\}$  for layers 0-5; $T_2$ for layers 6-11\} ; $\theta_{w}^{3} = \{0.1 \sigma$ for $\mathcal{P}_{1}$; $0.1 T_1$ for ATT; $0.2 T_2$ for FC\}; $\theta_{w}^{5} = \{\max\{T_1, T_2\}$  for layers 0-4; $T_2$ for layers 5-23\}}
\label{tab: pruning_FC}
\end{table}
Our method obtains WER of 23.32, 23.94 and 15.86, on Net-1, 2, 3, respectively, with SGL regularization before pruning. We conduct more experiments with various combinations of $T_1, T_2, T_3$ to set $\theta_{w}$ for the Net-1,2, 3 to provide the tradeoff between WER and pruning $\%$. A few experimental results are reported in Table \ref{tab: pruning_FC}. Net-1 gives lowest WER of $24.84$ with 35.4\% reduction in parameters. The "lowest WER" refers to the smallest WER obtained among all the experiments conducted for a model. The lowest WER reported for Net-2 is 32.86 with 32.2\% reduction in parameters. This indicates that all layers must not be regularized as non-FC layers capture significant information about the ASR learning. Hence, we refrain from performing regularization on all layers of Whisper-medium, but rather only on FC layers. Net-3 gives lowest WER of 19.58 when being pruned by $31\%$ of parameters. All our models substantially outperform the baseline IMP-based method both in terms of pruning \% and WER. Specifically, Net-1 prunes 18.7\% more parameters with a WER reduction of 12.31 when compared to IMP on Whisper-small. Similarly, on Whisper-medium, Net-3 achieves a WER reduction by 10.91 with 10.71\% more pruning. Further, in all our three networks, one can achieve a greater pruning $\%$ of upto $44.9$, $38.4$, and $51.5$ respectively, provided the increase in WER is tolerable for the ASR application considered. In all experiments, WERs shown are computed using our proposed custom Hindi normalizer. However, we also compare with other normalizers in Table \ref{tab: compare_normalizer}.
\begin{table}[h]
    \centering
    \begin{tabular}{|p{1.0cm}|c|c|c|c|}
    \hline
    \multirow{2}{*}{\textbf{Model}} & \multicolumn{2}{c|}{\textbf{Before pruning}} & \multicolumn{2}{c|}{\textbf{After pruning}} \\
    \cline{2-5}
    & Memory & FLOPs & Memory & FLOPs \\
    \hline
    Net-1 & 536.62 & 1.65e16 & 460.15 & 1.34e16 \\
    \hline
    Net-2 & 536.62 & 1.65e16 & 413.65 & 1.18e16 \\
    \hline
    Net-3 & 1557.11 & 4.95e16 & 1318.97 & 4.11e16 \\
    \hline
    \end{tabular}
    \caption{Memory costs (in MB) and FLOPs}
    \label{tab: mem_compute_cost}
\end{table}
From Table \ref{tab: mem_compute_cost}, we observe that memory consumption is reduced by 14.25\%, 22.91\%, 15.29\%, and computation costs (FLOPs) are decreased by 18.5\%, 28.31\%, 16.95\% in Net-1,2,3, respectively, for the lowest WER reported. The FLOPs are computed according to the methodology outlined in \cite{hoffmann2022training}.

In order to overcome shortcomings in existing Whisper and Indic normalizers, our proposed normalizer: (1) preserves diacritics except nasal marks, expands conjunct characters into constituent consonants without change in meaning; (2) removes punctuation but retains numerals and spacing to align model output and reference text; (3) maps common morphological variants, e.g., {\dn btAie}, {\dn btAao} (“tell”), to canonical forms; (4) normalizes acoustically similar vowels, e.g., ({\dn i} (IPA i), {\dn I} (IPA i:), to a standard representation for robustness to audio distortion. 
Its comparison with existing normalizers on WER evaluations is given in Table \ref{tab: compare_normalizer}.
\begin{table}[h]
    \centering
    \begin{tabular}{|l|l|c|c|}
    \hline
    \textbf{Model} & \textbf{Whisper} & \textbf{Indic} & \textbf{Ours} \\
    \hline
    VWM & 10.95 & 19.74 & 15.71 \\
    \hline
    \cite{bhogale2023vistaar} & 7.99 & 23.46 & 16.64\\
    \hline
    Net-3 & 10.98 & 20.03 & 15.86\\
    \hline
    VWS & 16.44 & 27.5 & 22.88\\
    \hline
    Net-1 & 16.53 & 28.09 & 23.32 \\
   \hline
    Net-2 & 17.04 & 28.57 & 23.94 \\
    \hline
    \end{tabular}
    \caption{WER comparison across different text normalizers before pruning}
    \label{tab: compare_normalizer}
\end{table}
VWS, VWM refer to Whisper-small, Whisper-medium, respectively, when fine-tuned without any regularization. Net-3 (15.86) has improved upon the state-of-the-art IndicWhisper (16.64). Note that IndicWhisper \cite{bhogale2023vistaar} was fine-tuned with 2150 hours of Hindi data, while we use only 13 hours of data. Net-1 maintains its WER (24.84 after pruning), comparable to its full model (23.32), and VWS (22.88).  Net-3 (19.58), after pruning, is also fairly competitive with full IndicWhisper (16.64). In Table \ref{tab: compare_normalizer} (columns ‘Whisper’, ‘Indic’, ‘Ours’), WERs obtained with our normalizer lie between WERs with Whisper and Indic, showing it balances the two extremes.
\section{CONCLUSION}\label{sec:conclusion}
In this paper, we propose a framework TSPAR to obtain compact fine-tuned versions of Whisper. TSPAR applies SGL loss regularization to induce both column-wise structured sparsity and element-wise unstructured sparsity. Structured sparsity gives us a significant reduction in FLOPs and model weights. TSPAR then performs a two-pass pruning adaptive to weight statistics, exploiting the induced sparsity.
As a result, we outperform the state-of-the-art IMP method both in pruning and WER performance, making our models suitable for resource constrained edge device applications.
We also design our custom Hindi text normalizer balancing the over-simplification by Whisper and over-preservation rules by Indic normalizers. 
A future direction is to leverage our TSPAR method to design low-complex variants of other state-of-the-art models under resource constraints.

\vfill\pagebreak
\bibliographystyle{IEEEbib}
\bibliography{refs_ft}

\end{document}